\documentclass[10pt,twocolumn,letterpaper]{article}

\usepackage{cvpr}              

\usepackage{graphicx}
\usepackage{amsmath}
\usepackage{amssymb}
\usepackage{booktabs}
\usepackage{multirow}
\usepackage{bbding}
\usepackage{color}
\usepackage[accsupp]{axessibility}  

\usepackage[pagebackref,breaklinks,colorlinks]{hyperref}

\usepackage[capitalize]{cleveref}
\crefname{section}{Sec.}{Secs.}
\Crefname{section}{Section}{Sections}
\Crefname{table}{Table}{Tables}
\crefname{table}{Tab.}{Tabs.}

\def\cvprPaperID{80} 
\def\confName{CVPR}
\def\confYear{2022}

\begin{document}

\title{Blueprint Separable Residual Network for Efficient Image Super-Resolution}

\author{
Zheyuan Li${^{1 *}}$ 
Yingqi Liu${^{1 *}}$ 
Xiangyu Chen${^{1,2 \dagger}}$ 
Haoming Cai${^1}$ 
Jinjin Gu${^{3,4}}$ 
Yu Qiao${^{1,3}}$ 
Chao Dong${^{1,3}}$ \\
$^{1}$ShenZhen Key Lab of Computer Vision and Pattern Recognition, SIAT-SenseTime Joint Lab,\\
Shenzhen Institutes of Advanced Technology, Chinese Academy of Sciences\\
${^2}$University of Macau 
${^3}$Shanghai AI Laboratory, Shanghai, China 
${^4}$The University of Sydney \\
{\tt\small \{zy.li3, yq.liu3, yu.qiao, chao.dong\}@siat.ac.cn, chxy95@gmail.com} \\
{\tt\small haomingcai@link.cuhk.edu.cn, jinjin.gu@sydney.edu.au}
}

\maketitle

\renewcommand{\thefootnote}{\fnsymbol{footnote}}
\footnotetext[1]{~indicates contribute equally. $^\dag$~Corresponding author.}
\renewcommand{\thefootnote}{1}

\begin{abstract}
Recent advances in single image super-resolution (SISR) have achieved extraordinary performance, but the computational cost is too heavy to apply in edge devices. To alleviate this problem, many novel and effective solutions have been proposed. Convolutional neural network (CNN) with the attention mechanism has attracted increasing attention due to its efficiency and effectiveness. However, there is still redundancy in the convolution operation. In this paper, we propose Blueprint Separable Residual Network (BSRN) containing two efficient designs. One is the usage of blueprint separable convolution (BSConv), which takes place of the redundant convolution operation. The other is to enhance the model ability by introducing more effective attention modules. The experimental results show that BSRN achieves state-of-the-art performance among existing efficient SR methods. Moreover, a smaller variant of our model BSRN-S won the first place in model complexity track of NTIRE 2022 Efficient SR Challenge. The code is available at \url{https://github.com/xiaom233/BSRN}.
\end{abstract}

\section{Introduction}

\begin{figure}[]
  \centering
    \includegraphics[width=1\linewidth]{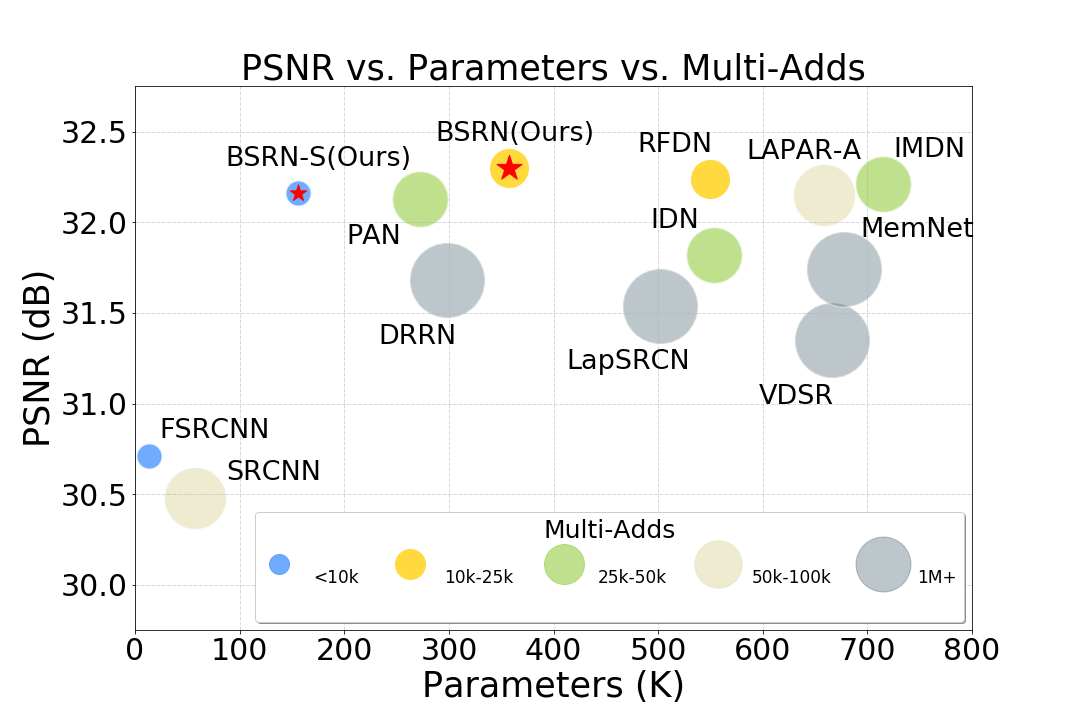}
  \caption{Performance and model complexity comparison on Set5 dataset for upscaling factor $\times4$.}
  \label{fig:scatter}
\end{figure}

Single image super-resolution (SR) is a fundamental task in the computer vision field. It aims at reconstructing a visual-pleasing high-resolution (HR) image from the corresponding low-resolution (LR) observation.
In recent years, the general paradigm has gradually shifted from model-based solutions to deep learning methods~\cite{dong2014learning,kim2016accurate,lim2017enhanced,zhang2018image,chen2021attention,liang2021swinir}.
These SR networks have greatly improved the quality of restored images.
Their success can be partially attributed to the large model capacity and intensive computation. 
However, these properties could largely limit their application in real-world scenarios that prefer efficiency or require real-time implementation. 
Many lightweight SR networks have been proposed to address the inefficient issue.
These approaches use different strategies to achieve high efficiency, including parameter sharing strategy~\cite{kim2016deeply,tai2017image}, cascading network with grouped convolution~\cite{ahn2018fast}, information or feature distillation mechanisms~\cite{hui2018fast,hui2019lightweight,rfdn} and attention mechanisms~\cite{zhao2020efficient,chen2021attention}.
While they have applied compact architectures and improved mapping efficiency, there still exists redundancy in convolution operations. 
We can build more efficient SR networks by reducing redundant computations and exploiting more effective modules.

In this paper, we propose a new lightweight SR network, namely Blueprint Separable Residual Network (BSRN), which improves the network's efficiency from two perspectives --- optimizing the convolutional operations and introducing effective attention modules.
First, as the name suggests, BSRN reduces redundancies by using blueprint separation convolutions (BSConv) \cite{haase2020rethinking} to construct the basic building blocks. 
BSConv is an improved variant of the original depth-wise separable convolution (DSConv) \cite{howard2017mobilenets}, which better exploits intra-kernel correlations for an efficient separation~\cite{haase2020rethinking}.
%
%
%
Our work shows that BSConv is beneficial for efficient SR.
%
%
%
%
Second, appropriate attention modules~\cite{hui2019lightweight,rfdn,zhao2020efficient,rfan} have been shown to improve the performance of efficient SR networks. 
Inspired by these works, we also introduce two effective attention modules, enhanced spatial attention (ESA)~\cite{rfan} and contrast-aware channel attention (CCA)~\cite{hui2019lightweight}, to enhance the model ability.
The proposed BSRN method achieves state-of-the-art performance among existing efficiency-oriented SR networks, as shown in~\cref{fig:scatter}.
We took a variant of our method BSRN-S to participate in the NTIRE 2022 Efficient SR Challenge and won first place in the model complexity track~\cite{li2022ntire}.
%

The main contributions of this paper are:
\begin{itemize}
\vspace{-6pt}
\setlength{\itemsep}{0pt}
\setlength{\parsep}{0pt}
\setlength{\parskip}{0pt}
\item We introduce BSConv to construct the basic building block and show its effectiveness for SR.
\item We utilize two effective attention modules with limited extra computation to enhance the model ability.
\item The proposed BSRN, which integrates BSConv and effective attention modules demonstrates superior performance for efficient SR.
\end{itemize}

\begin{figure*}[h!]
  \centering
    \includegraphics[width=17cm]{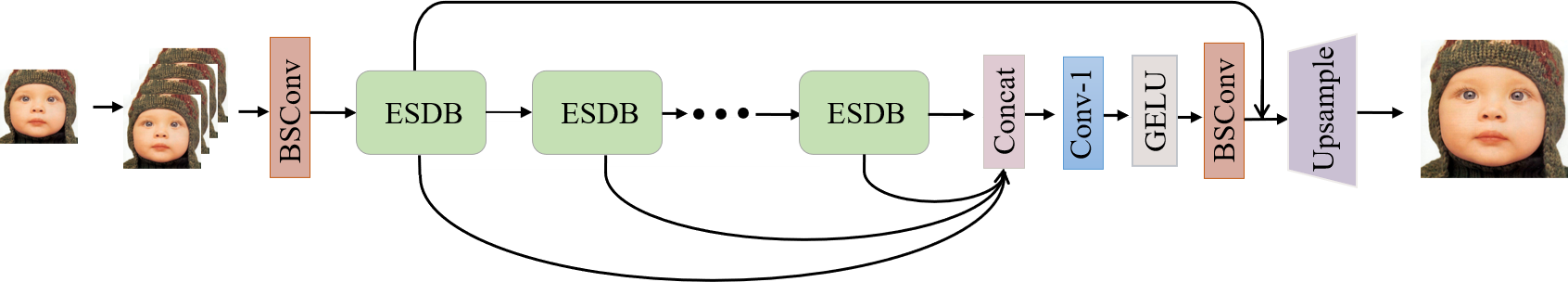}
  \caption{The architecture of Blueprint Separable Residual Network (BSRN).}
  \label{fig:BSRN_arch}
\end{figure*}

\section{Related Work}
\subsection{Deep Networks for SR}
With the fast development of deep learning techniques, increasing remarkable progress has been made for the SR task. Since Dong \etal~\cite{dong2014learning} proposed the pioneering work SRCNN with a three-layer convolutional neural network and significantly outperforms the conventional methods, a series of methods \cite{kim2016accurate,zhang2018residual,zhang2018residual,zhang2018image,liang2021swinir} have been proposed to improve the SR model. For example, Kim \etal~\cite{kim2016accurate} proved that a deeper network can get better performance by increasing the depth of the network to 20. Zhang \etal~\cite{zhang2018residual} introduced dense connection into the network to further enhance the representative ability of the model. \cite{zhang2018image} introduced the channel-wise attention mechanism to utilize the global statistics for better performance. Liang \etal~\cite{liang2021swinir} proposed a Transformer architecture for image restoration based on the Swin Transformer~\cite{liu2021swin}, which achieves a significant improvement and refreshes the state-of-the-art performance. Although the abovementioned approaches make great progress in performance, most of them bring high computational costs, which prompts researchers to develop more efficient methods for the SR task.

\subsection{CNN Model Compression and Acceleration}
During the past few years, tremendous progress has been made in the area of model compression and acceleration. In general, these techniques can be divided into four categories~\cite{cheng2017survey}: parameter pruning and quantization, low-rank factorization, knowledge distillation, and transferred/compact convolutional filters. For parameter pruning~\cite{hanson1988comparing, srinivas2015data, lebedev2016fast, li2020dhp} and quantization methods~\cite{vanhoucke2011improving, han2015deep, courbariaux2015binaryconnect}, they aim to explore the redundancy of the model architecture and try to remove or reduce the redundant parameters. 
The low-rank factorization approaches~\cite{rigamonti2013learning, denil2013predicting} use matrix/tensor decomposition to estimate the more informative representation of the networks. Knowledge distillation methods\cite{44873, korattikara2015bayesian} aim to generate more compact student models from a larger network by learning the distributions of teacher models. The methods based on transferred/compact convolutional filters design~\cite{iandola2016squeezenet, howard2017mobilenets,Sandler_2018_CVPR,zhang2018shufflenet,li2019learning,howard2019searching,haase2020rethinking} aim to devise special structural convolutional filters to reduce the model parameters and save storage/computation.

\subsection{Efficient SR Models}
Most of the current models for SR often introduce lots of computational costs when bringing performance improvements, which restricts the practical application of these methods. Thus, many works have been proposed to design more efficient models for the task~\cite{dong2016accelerating, kim2016deeply,li2021heterogeneity,tai2017image,song2020efficient,ahn2018fast,hui2019lightweight,zhao2020efficient,rfdn,wu2021trilevel}. For instance, \cite{dong2016accelerating} directly used the original LR images as input instead of the pre-upsampled ones and placed a deconvolution at the end of the network to save computation.
\cite{ahn2018fast} uses group convolution to reduce the computation of the standard convolution. 
\cite{hui2018fast} proposed an information distillation network (IDN) that explicitly splits features and then processes them separately.
\cite{hui2019lightweight} designed an information multi-distillation block that split features and refine them step by step to reduce the computation.
\cite{zhao2020efficient} introduced pixel attention and self-calibrated convolution to use fewer parameters to achieve competitive performance. 
\cite{rfdn} proposed residual feature distillation block by improving the information distillation mechanism and won the championship of AIM 2020 Efficient SR Challenge~\cite{zhang2020aim}.

\section{Method}
\subsection{Network Architecture}
\begin{figure*}[h!]
  \centering
    \includegraphics[width=16.5cm]{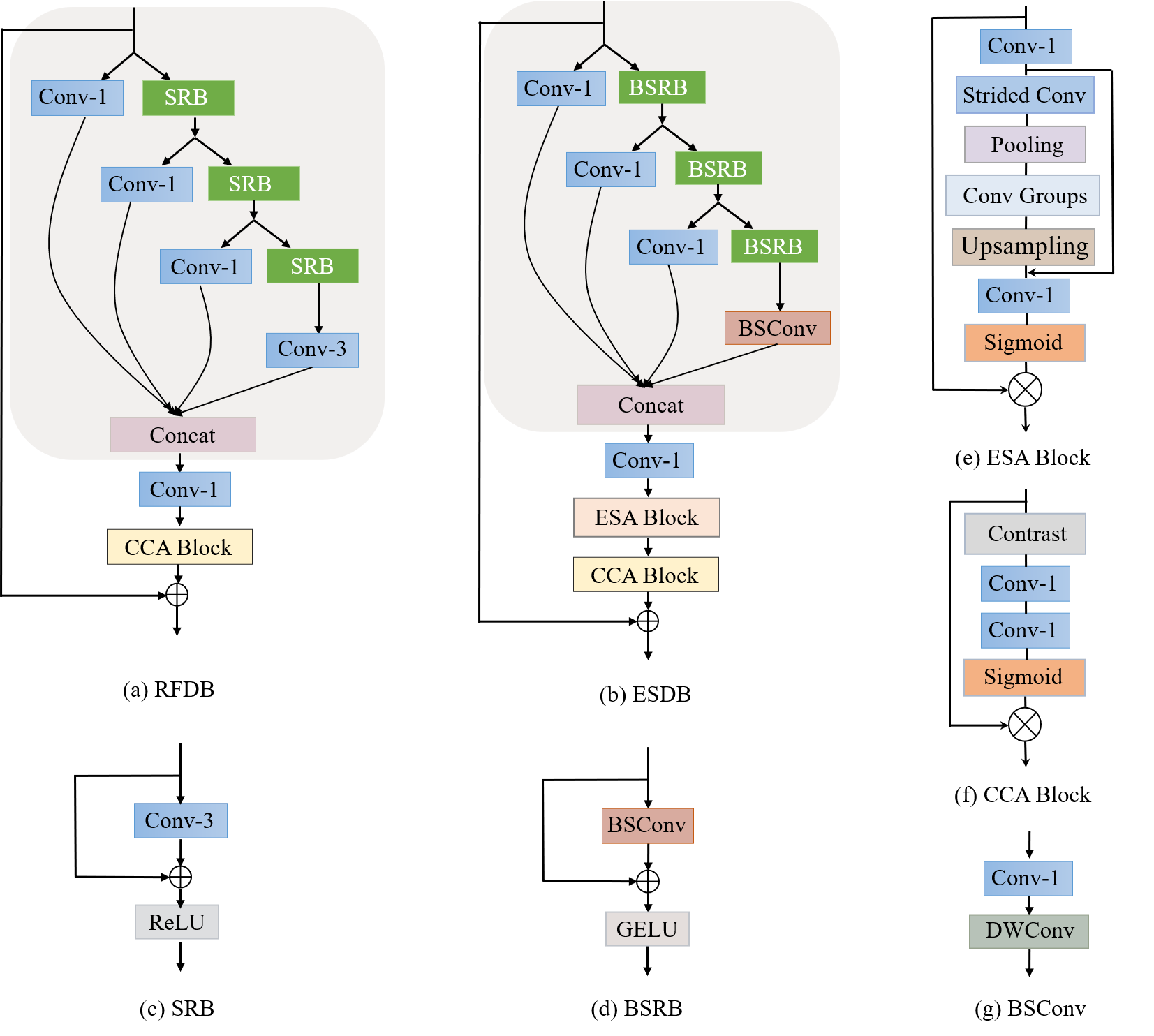}
  \caption{(a) The architecture of RFDB. (b) The architecture of the proposed ESDB. (c) The architecture of ESA block. (d) The architecture of channel weighting. (e) The architecture of SRB in RFDB. (f) The architecture of the proposed BSRB in ESDB. (g) The architecture of BSConv,   consists of a $1\times1$ convolution layer and a depth-wise convolution layer.}
  \label{fig:modules}
\end{figure*}

The overall architecture of our method BSRN is shown in \cref{fig:BSRN_arch}. It is inherited from the structure of RFDN~\cite{rfdn}, which is the champion solution of AIM 2020 Challenge on Efficient Super-Resolution.
It consists of four stages: shallow feature extraction, deep feature extraction, multi-layer feature fusion and reconstruction.
Let us denote $I_{LR}$ and $I_{SR}$ as the input and output image.
In pre-processing, the input image is first replicated n times. Then we concatenate these images together as
\begin{equation}
I_{LR}^n = Concat_n(I_{LR}),
\end{equation}
where $Concat(\cdot)$ denotes the concatenation operation along the channel dimension, and $n$ is the number of $I_{LR}$ to be concatenated.
The next shallow feature extraction part maps the input image to a higher dimensional feature space as 
\begin{equation}
F_0 = H_{SF}(I_{LR}^n),
\end{equation}
where $H_{SF}(\cdot)$ denotes the module of shallow feature extraction. To be specific, we use a BSConv~\cite{haase2020rethinking} to achieve shallow feature extraction. The specific architecture of BSConv is depicted in \cref{fig:modules} (g), which consists of a $1\times1$ convolution and a depth-wise convolution. $F_0$ is then used for the deep feature extraction by a stack of ESDBs, which gradually refine the extracted features. This process can be formulated as
\begin{equation}
F_k=H_k(F_{k-1}),k=1,...,n,
\end{equation}
where $H_k(\cdot)$ denotes the $k$-th ESDB. $F_{k-1}$ and $F_k$ represent the input feature and output feature of the $k$-th ESDB, respectively. To fully utilize features from all depths, features generated at different depths are fused and mapped by a $1\times1$ convolution and a GELU\cite{hendrycks2016gaussian} activation. Then, a BSConv is used to refine features. The multi-layer feature fusion is formulated as
\begin{equation}
F_{fused}=H_{fusion}(Concat(F_1,...F_{k-1})),
\end{equation}
where $H_{fusion}(\cdot)$ represents the fusion module and $F_{fused}$ is the aggregated feature. To take advantage of residual learning, a long skip connection is involved. The reconstruction stage is formulated as
\begin{equation}
I_{SR}=H_{BSRN}(I_{LR}^i)=H_{rec}(F_{fusion}+F_0),
\end{equation}
where $H_{rec}(\cdot)$ denotes the reconstruction module, which consists of a $3\times3$ standard convolution layer and a pixel-shuffle operation~\cite{pixelshuffle}.
$L_1$ loss function is exploited to optimize the model, which can be formulated as 
\begin{equation}
  L_1=\Vert I_{SR}-I_{HR} \Vert_1.
\end{equation}

\subsection{Efficient Separable Distillation Block}
\label{section:esdb}
Inspired by the RFDB in RFDN~\cite{rfdn}, we design the efficient separable distillation block (ESDB) that is similar to RFDB in structure but more efficient.
The overall architecture of ESDB is shown in \cref{fig:modules} (b). An ESDB generally consists of 3 stages: feature distillation, feature condensation and feature enhancement. In the first stage, for an input feature $F_{in}$, the feature distillation can be formulated as 
\begin{equation}
    \begin{split}
        F_{distilled\_1},F_{coarse\_1}&=DL_1(F_{in}),RL_1(F_{in}),\\
        F_{distilled\_2},F_{coarse\_2}&=DL_2(F_{coarse\_1}),RL_2(F_{coarse\_1}),\\
        F_{distilled\_3},F_{coarse\_3}&=DL_3(F_{coarse\_2}),RL_3(F_{coarse\_2}),\\
        F_{distilled\_4}&=DL_4(F_{coarse\_3}),
    \end{split}
\end{equation}
where $DL$ denotes the distillation layer that generate distilled features, and $RL$ denotes the refinement layer that further refines the coarse feature step by step.
In the feature condensation stage, the distilled features $F_{distilled\_1}$, $F_{distilled\_2}$, $F_{distilled\_3}$, $F_{distilled\_4}$ are concatenated together and then condensed by a $1\times1$ convolution as
\begin{equation}
F_{condensed}=H_{linear}(Concat(F_{distilled\_1},...,F_{distilled\_4})),
\end{equation}
where $F_{condensed}$ is the condensed feature, $H_{linear}(\cdot)$ denotes the $1\times1$ convolution layer. For the last stage, to enhance the representational ability of the model while keeping efficiency, we introduce a lightweight enhanced spatial attention (ESA) block~\cite{rfan} and a contrast-aware channel attention (CCA) block~\cite{hui2019lightweight} as 
\begin{equation}
F_{enhanced}=H_{CCA}(H_{ESA}(F_{condensed})),
\end{equation}
where $F_{enhanced}$ is the enhanced feature, $H_{ESA}(\cdot)$ and $H_{CCA}(\cdot)$ denote the ESA and CCA modules that have been shown to enhance the model ability effectively~\cite{rfan,hui2019lightweight} from the spatial and channel-wise perspective, respectively.


\begin{table*}[htb!]
\centering
\caption{Quantitative comparison of different convolution decomposition approaches. BSConvU is exploited in our method as BSConv.}
\label{table:ablation_conv}
\resizebox{\linewidth}{!}{
\begin{tabular}{c|cc|cc|cc|cc|cc|cc}
\toprule
\multirow{2}{*}{Method} & \multirow{2}{*}{Params[K]}  & \multirow{2}{*}{Multi-Adds[G]}  & \multicolumn{2}{c|}{Set5} & \multicolumn{2}{c|}{Set14} & \multicolumn{2}{c|}{B100} & \multicolumn{2}{c|}{Urban100} & \multicolumn{2}{c}{Manga109} \\ \cline{4-13} 
                        &            &         & PSNR        & SSIM        & PSNR         & SSIM        & PSNR        & SSIM        & PSNR          & SSIM          & PSNR          & SSIM         \\ \hline
    RFDN                &      433K  & 23.9   & 32.04       & 0.8934      & 28.52        & 0.7799      & 27.53       & 0.7344      & 25.92         & 0.7810        & 30.30         & 0.9063       \\
    RFDN-DSConv         &      123K  & 6.9    & 31.95       & 0.8910      & 28.40        & 0.7772      & 27.45       & 0.7318      & 25.64         & 0.7726        & 29.84         & 0.9010       \\
    RFDN-BSConvS        &      122K  & 6.8    & 31.94       & 0.8917      & 28.44        & 0.7777      & 27.48       & 0.7322      & 25.70         & 0.7731        & 30.03         & 0.9027       \\
    RFDN-BSConvU        &      124K  & 6.8    & 31.99       & 0.8921      & 28.46        & 0.7783      & 27.47       & 0.7324      & 25.72         & 0.7742        & 29.99         & 0.9022       \\ 
    \bottomrule
\end{tabular}}
\end{table*}

\begin{table*}[t!]
\centering
\caption{Ablation study of ESA and CCA.}
\label{table:ablation_module}
\resizebox{\linewidth}{!}{
\begin{tabular}{c|cc|cc|cc|cc|cc|cc}
\toprule
\multirow{2}{*}{Method} & \multirow{2}{*}{Params[K]}  & \multirow{2}{*}{Multi-Adds[G]} & \multicolumn{2}{c|}{Set5} & \multicolumn{2}{c|}{Set14} & \multicolumn{2}{c|}{B100} & \multicolumn{2}{c|}{Urban100} & \multicolumn{2}{c}{Manga109} \\ \cline{4-13} 
                 &        &             & PSNR        & SSIM        & PSNR         & SSIM        & PSNR        & SSIM        & PSNR          & SSIM          & PSNR          & SSIM         \\ \hline
    BSRN-woESA     & 320    &   18.2      &  32.14      & 0.8943      & 28.56        & 0.7807      & 27.56       & 0.7352      & 25.97         & 0.7816        & 30.39         &  0.9071      \\ 
    BSRN-woCCA    & 348    &   19.4      &  32.20      & 0.8947      & 28.65        & 0.7824      & 27.60       & 0.7368      & 26.05         & 0.7854        & 30.53         &  0.9087      \\ 
    BSRN & 352    &   19.4      &  32.25      & 0.8956      & 28.62        & 0.7822      & 27.60       & 0.7367      & 26.10         & 0.7864        & 30.58         &  0.9093      \\ \bottomrule
\end{tabular}}
\end{table*}

\begin{table*}[htb!]
\centering
\caption{Quantitative comparison of different activation functions.}
\label{table:activation}
\resizebox{\linewidth}{!}{
\begin{tabular}{c|c|cc|cc|cc|cc|cc}
\toprule
\multirow{2}{*}{Method} & \multirow{2}{*}{DIV2K\_val} & \multicolumn{2}{c|}{Set5} & \multicolumn{2}{c|}{Set14} & \multicolumn{2}{c|}{B100} & \multicolumn{2}{c|}{Urban100} & \multicolumn{2}{c}{Manga109} \\ \cline{3-12} 
              &                         & PSNR        & SSIM        & PSNR         & SSIM        & PSNR        & SSIM        & PSNR          & SSIM          & PSNR          & SSIM         \\ \hline
    ReLU      & 28.95                   &  32.15      & 0.8943      & 28.59        & 0.7815      & 27.57       & 0.7358      & 26.02         & 0.7836        & 30.49         &  0.9082      \\
    LeakyReLU & 28.97                   &  32.24      & 0.8953      & 28.58        & 0.7817      & 27.58       & 0.7361      & 26.07         & 0.7854        & 30.55         &  0.9092      \\
    h-swish   &  28.99                  &  32.22      & 0.8952      & 28.61        & 0.7825      & 27.59       & 0.7363      & 26.07         & 0.7851        & 30.50         &  0.9083      \\
    GELU      & 29.00                   &  32.25      & 0.8956      & 28.62        & 0.7822      & 27.60       & 0.7367      & 26.10         & 0.7864        & 30.58         &  0.9093      \\ \bottomrule
\end{tabular}}
\end{table*}

\begin{table*}[t!]
\centering
\caption{Quantitative comparison of two BSRN variants with RFDN. BSRN-1 has the same depth and width as RFDN, while BSRN-2 has similar computational complexity to RFDN.}
\label{table:BSRN_effectivenes}
\resizebox{\linewidth}{!}{
\begin{tabular}{c|cc|cc|cc|cc|cc|cc}
\toprule
\multirow{2}{*}{Method} & \multirow{2}{*}{Params[K]} & \multirow{2}{*}{Multi-Adds[G]} & \multicolumn{2}{c|}{Set5} & \multicolumn{2}{c|}{Set14} & \multicolumn{2}{c|}{B100} & \multicolumn{2}{c|}{Urban100} & \multicolumn{2}{c}{Manga109} \\ \cline{4-13} 
                 &        &                 & PSNR        & SSIM        & PSNR         & SSIM        & PSNR        & SSIM        & PSNR          & SSIM          & PSNR          & SSIM         \\ \hline
    RFDN         & 443    &     23.9        &  32.04      & 0.8934      & 28.52        & 0.7799      & 27.53       & 0.7344      & 25.92         & 0.7810        & 30.30         &  0.9063      \\
    BSRN-1       & 209    &     11.5        &  32.14      & 0.8942      & 28.57        & 0.7811      & 27.55       & 0.7352      & 25.95         & 0.7815        & 30.35         &  0.9068      \\ 
    BSRN-2       & 438    &     24.2        &  32.22      & 0.8954      & 28.62        & 0.7827      & 27.60       & 0.7369      & 26.08         & 0.7855        & 30.61         &  0.9096      \\ \bottomrule
\end{tabular}}
\end{table*}

\textbf{Blueprint Shallow Residual Block (BSRB).}
A basic module of ESDB is BSRB, as shown in~\cref{fig:modules} (d), which consists of a BSConv, an identity connection and an activation unit. Specifically, we use GELU~\cite{GELU} as the activation function. BSConv factorizes a standard convolution into a point-wise $1\times1$ convolution and a depth-wise convolution, as depicted in~\cref{fig:modules} (g). It is an inverse version of the depth-wise separable convolution (DSConv)\cite{howard2017mobilenets}. \cite{bsconv} shows that BSConv performs better in many cases for efficient separation of the standard convolution, thus we exploit it in our model. For the activation unit, GELU~\cite{GELU} gradually becomes the first choice in recent works~\cite{devlin2018bert,radford2019language,liu2021swin, liang2021swinir}, which can be seen as a smoother variant of ReLU. In our method, we also find that GELU performs better than the commonly used ReLU\cite{nair2010rectified} and LeakyReLU\cite{maas2013rectifier}.

\textbf{Attention modules of ESA and CCA.} Since the effectiveness of ESA and CCA has been proven \cite{rfan,hui2019lightweight,rfdn}, we introduce the two modules into our approach. The specific architecture of the ESA block is shown in \cref{fig:modules} (f). It starts with a $1\times 1$ convolutional layer to reduce the channel dimensions of the input feature. Then the block uses a strided convolution and a strided max-pooling layer to reduce the spatial size. Following a group of convolutions to extract the feature, an interpolation-based up-sampling is performed to recover the spatial size. Note that the convolutions in our ESA are also BSConvs for better efficiency different from the original version~\cite{rfan}. Combined with a residual connection, the features are further processed by a $1\times 1$ convolutional layer to restore the channel size. Finally, the attention matrix is generated via a Sigmoid function and multiplied by the original input feature. A CCA block is added after the ESA block shown in~\cref{fig:modules} (f), which is an improved version of the channel attention module proposed for the SR task~\cite{hui2019lightweight}. Different from the conventional channel attention calculated using the mean of each channel-wise feature, CCA utilizes the contrast information including the mean and the summation of standard deviation to calculate the channel attention weights. 

\section{Experiments}

\begin{table*}[!t]
\centering
\caption{Quantitative comparison with state-of-the-art methods on benchmark datasets. The best and second-best performance are in \textcolor{red}{red} and \textcolor{blue}{blue} colors, respectively. 'Multi-Adds' is calculated with a $1280\times720$ GT image.}
\label{table:comparison_SOTA}
\resizebox{\linewidth}{!}{
\begin{tabular}{l|c|cc|ll|ll|ll|ll|ll}
\toprule
\multicolumn{1}{c|}{\multirow{2}{*}{Method}} & \multirow{2}{*}{Scale} & \multirow{2}{*}{Params[K]} & \multirow{2}{*}{Multi-Adds[G]}          & \multicolumn{2}{c|}{Set5}         & \multicolumn{2}{c|}{Set14}        & \multicolumn{2}{c|}{BSD100}       & \multicolumn{2}{c|}{Urban100}     & \multicolumn{2}{c}{Manga109}     \\ \cline{4-13} 
\multicolumn{1}{c|}{}                        &                        &            &            & \multicolumn{2}{l|}{PSNR/SSIM}    & \multicolumn{2}{l|}{PSNR/SSIM}    & \multicolumn{2}{l|}{PSNR/SSIM}    & \multicolumn{2}{l|}{PSNR/SSIM}    & \multicolumn{2}{l}{PSNR/SSIM}    \\ \hline
Bicubic                                       & \multirow{16}{*}{$\times 2$}   & -         &  -          & \multicolumn{2}{l|}{33.66/0.9299} & \multicolumn{2}{l|}{30.24/0.8688} & \multicolumn{2}{l|}{29.56/0.8431} & \multicolumn{2}{l|}{26.88/0.8403} & \multicolumn{2}{l}{30.80/0.9339} \\
SRCNN\cite{dong2014learning}                  &                        & 8        &  52.7      & \multicolumn{2}{l|}{36.66/0.9542} & \multicolumn{2}{l|}{32.45/0.9067} & \multicolumn{2}{l|}{31.36/0.8879} & \multicolumn{2}{l|}{29.50/0.8946} & \multicolumn{2}{l}{35.60/0.9663} \\
FSRCNN\cite{dong2016accelerating}             &                        & 13       &  6.0      & \multicolumn{2}{l|}{37.00/0.9558} & \multicolumn{2}{l|}{32.63/0.9088} & \multicolumn{2}{l|}{31.53/0.8920} & \multicolumn{2}{l|}{29.88/0.9020} & \multicolumn{2}{l}{36.67/0.9710} \\
VDSR\cite{kim2016accurate}                    &                        & 666      &  612.6     & \multicolumn{2}{l|}{37.53/0.9587} & \multicolumn{2}{l|}{33.03/0.9124} & \multicolumn{2}{l|}{31.90/0.8960} & \multicolumn{2}{l|}{30.76/0.9140} & \multicolumn{2}{l}{37.22/0.9750} \\
LapSRN\cite{lai2017deep}                      &                        & 251      &  29.9      & \multicolumn{2}{l|}{37.52/0.9591} & \multicolumn{2}{l|}{32.99/0.9124} & \multicolumn{2}{l|}{31.80/0.8952} & \multicolumn{2}{l|}{30.41/0.9103} & \multicolumn{2}{l}{37.27/0.9740} \\
DRRN\cite{tai2017image}                       &                        & 298      &  6,796.9   & \multicolumn{2}{l|}{37.74/0.9591} & \multicolumn{2}{l|}{33.23/0.9136} & \multicolumn{2}{l|}{32.05/0.8973} & \multicolumn{2}{l|}{31.23/0.9188} & \multicolumn{2}{l}{37.88/0.9749} \\
MemNet\cite{tai2017memnet}                    &                        & 678      &   2,662.4  & \multicolumn{2}{l|}{37.78/0.9597} & \multicolumn{2}{l|}{33.28/0.9142} & \multicolumn{2}{l|}{32.08/0.8978} & \multicolumn{2}{l|}{31.31/0.9195} & \multicolumn{2}{l}{37.72/0.9740} \\
IDN\cite{hui2018fast}                         &                        & 553      &  124.6     & \multicolumn{2}{l|}{37.83/0.9600} & \multicolumn{2}{l|}{33.30/0.9148} & \multicolumn{2}{l|}{32.08/0.8985} & \multicolumn{2}{l|}{31.27/0.9196} & \multicolumn{2}{l}{38.01/0.9749} \\
CARN\cite{ahn2018fast}                        &                        & 1592     &  222.8     & \multicolumn{2}{l|}{37.76/0.9590} & \multicolumn{2}{l|}{33.52/0.9166} & \multicolumn{2}{l|}{32.09/0.8978} & \multicolumn{2}{l|}{31.92/0.9256} & \multicolumn{2}{l}{38.36/0.9765} \\
IMDN\cite{hui2019lightweight}                 &                        & 694      &  158.8     & \multicolumn{2}{l|}{38.00/0.9605} & \multicolumn{2}{l|}{33.63/0.9177} & \multicolumn{2}{l|}{32.19/0.8996} & \multicolumn{2}{l|}{\textcolor{blue}{32.17/0.9283}} & \multicolumn{2}{l}{\textcolor{blue}{38.88/0.9774}} \\
PAN\cite{zhao2020efficient}                   &                        & 261      &  70.5      & \multicolumn{2}{l|}{38.00/0.9605} & \multicolumn{2}{l|}{33.59/0.9181} & \multicolumn{2}{l|}{32.18/0.8997} & \multicolumn{2}{l|}{32.01/0.9273} & \multicolumn{2}{l}{38.70/0.9773} \\
LAPAR-A\cite{li2020lapar}                     &                        & 548      &   171.0    & \multicolumn{2}{l|}{38.01/0.9605} & \multicolumn{2}{l|}{33.62/0.9183} & \multicolumn{2}{l|}{\textcolor{blue}{32.19/0.8999}} & \multicolumn{2}{l|}{32.10/0.9283} & \multicolumn{2}{l}{38.67/0.9772} \\
RFDN\cite{rfdn}                    &                        & 534      &  95.0      & \multicolumn{2}{l|}{\textcolor{blue}{38.05/0.9606}} & \multicolumn{2}{l|}{\textcolor{blue}{33.68/0.9184}} & \multicolumn{2}{l|}{32.16/0.8994} & \multicolumn{2}{l|}{32.12/0.9278} & \multicolumn{2}{l}{38.88/0.9773} \\
BSRN(Ours)                                    &                        & 332      &  73.0      & \multicolumn{2}{l|}{\textcolor{red}{38.10/0.9610}} & \multicolumn{2}{l|}{\textcolor{red}{33.74/0.9193}} & \multicolumn{2}{l|}{\textcolor{red}{32.24/0.9006}} & \multicolumn{2}{l|}{\textcolor{red}{32.34/0.9303}} & \multicolumn{2}{l}{\textcolor{red}{39.14/0.9782}} \\ \hline
Bicubic                                       & \multirow{16}{*}{$\times 3$}   & -         &  -          & \multicolumn{2}{l|}{30.39/0.8682} & \multicolumn{2}{l|}{27.55/0.7742} & \multicolumn{2}{l|}{27.21/0.7385} & \multicolumn{2}{l|}{24.46/0.7349} & \multicolumn{2}{l}{26.95/0.8556} \\
SRCNN\cite{dong2014learning}                  &                        & 8        &  52.7      & \multicolumn{2}{l|}{32.75/0.9090} & \multicolumn{2}{l|}{29.30/0.8215} & \multicolumn{2}{l|}{28.41/0.7863} & \multicolumn{2}{l|}{26.24/0.7989} & \multicolumn{2}{l}{30.48/0.9117} \\
FSRCNN\cite{dong2016accelerating}             &                        & 13       &  5.0       & \multicolumn{2}{l|}{33.18/0.9140} & \multicolumn{2}{l|}{29.37/0.8240} & \multicolumn{2}{l|}{28.53/0.7910} & \multicolumn{2}{l|}{26.43/0.8080} & \multicolumn{2}{l}{31.10/0.9210} \\
VDSR\cite{kim2016accurate}                    &                        & 666      &  612.6     & \multicolumn{2}{l|}{33.66/0.9213} & \multicolumn{2}{l|}{29.77/0.8314} & \multicolumn{2}{l|}{28.82/0.7976} & \multicolumn{2}{l|}{27.14/0.8279} & \multicolumn{2}{l}{32.01/0.9340} \\
DRRN\cite{tai2017image}                       &                        & 298      &  6,796,9   & \multicolumn{2}{l|}{34.03/0.9244} & \multicolumn{2}{l|}{29.96/0.8349} & \multicolumn{2}{l|}{28.95/0.8004} & \multicolumn{2}{l|}{27.53/0.8378} & \multicolumn{2}{l}{32.71/0.9379} \\
MemNet\cite{tai2017memnet}                    &                        & 678      &  2,662.4   & \multicolumn{2}{l|}{34.09/0.9248} & \multicolumn{2}{l|}{30.00/0.8350} & \multicolumn{2}{l|}{28.96/0.8001} & \multicolumn{2}{l|}{27.56/0.8376} & \multicolumn{2}{l}{32.51/0.9369} \\
IDN\cite{hui2018fast}                         &                        & 553      &  56.3      & \multicolumn{2}{l|}{34.11/0.9253} & \multicolumn{2}{l|}{29.99/0.8354} & \multicolumn{2}{l|}{28.95/0.8013} & \multicolumn{2}{l|}{27.42/0.8359} & \multicolumn{2}{l}{32.71/0.9381} \\
CARN\cite{ahn2018fast}                        &                        & 1592     &  118.8     & \multicolumn{2}{l|}{34.29/0.9255} & \multicolumn{2}{l|}{30.29/0.8407} & \multicolumn{2}{l|}{29.06/0.8034} & \multicolumn{2}{l|}{28.06/0.8493} & \multicolumn{2}{l}{33.50/0.9440} \\
IMDN\cite{hui2019lightweight}                 &                        & 703      &  71.5      & \multicolumn{2}{l|}{34.36/0.9270} & \multicolumn{2}{l|}{30.32/0.8417} & \multicolumn{2}{l|}{29.09/0.8046} & \multicolumn{2}{l|}{28.17/0.8519} & \multicolumn{2}{l}{33.61/0.9445} \\
PAN\cite{zhao2020efficient}                   &                        & 261      &  39.0      & \multicolumn{2}{l|}{34.40/0.9271} & \multicolumn{2}{l|}{30.36/0.8423} & \multicolumn{2}{l|}{29.11/0.8050} & \multicolumn{2}{l|}{28.11/0.8511} & \multicolumn{2}{l}{33.61/0.9448} \\
LAPAR-A\cite{li2020lapar}                     &                        & 544      &   114.0    & \multicolumn{2}{l|}{34.36/0.9267} & \multicolumn{2}{l|}{\textcolor{blue}{30.34/0.8421}} & \multicolumn{2}{l|}{\textcolor{blue}{29.11/0.8054}} & \multicolumn{2}{l|}{28.15/0.8523} & \multicolumn{2}{l}{33.51/0.9441} \\
RFDN\cite{rfdn}                    &                        & 541      &  42.2      & \multicolumn{2}{l|}{\textcolor{blue}{34.41/0.9273}} & \multicolumn{2}{l|}{30.34/0.8420} & \multicolumn{2}{l|}{29.09/0.8050} & \multicolumn{2}{l|}{\textcolor{blue}{28.21}/\textcolor{blue}{0.8525}} & \multicolumn{2}{l}{\textcolor{blue}{33.67/0.9449}} \\
BSRN(Ours)                                  &                        & 340      &  33.3      & \multicolumn{2}{l|}{\textcolor{red}{34.46/0.9277}} & \multicolumn{2}{l|}{\textcolor{red}{30.47/0.8449}} & \multicolumn{2}{l|}{\textcolor{red}{29.18/0.8068}} & \multicolumn{2}{l|}{\textcolor{red}{28.39/0.8567}} & \multicolumn{2}{l}{\textcolor{red}{34.05/0.9471}} \\ \hline
Bicubic                                       & \multirow{16}{*}{$\times 4$}   & -         &  -          & \multicolumn{2}{l|}{28.42/0.8104} & \multicolumn{2}{l|}{26.00/0.7027} & \multicolumn{2}{l|}{25.96/0.6675} & \multicolumn{2}{l|}{23.14/0.6577} & \multicolumn{2}{l}{24.89/0.7866} \\
SRCNN\cite{dong2014learning}                  &                        & 8        &  52.7      & \multicolumn{2}{l|}{30.48/0.8626} & \multicolumn{2}{l|}{27.50/0.7513} & \multicolumn{2}{l|}{26.90/0.7101} & \multicolumn{2}{l|}{24.52/0.7221} & \multicolumn{2}{l}{27.58/0.8555} \\
FSRCNN\cite{dong2016accelerating}             &                        & 13       &  4.6       & \multicolumn{2}{l|}{30.72/0.8660} & \multicolumn{2}{l|}{27.61/0.7550} & \multicolumn{2}{l|}{26.98/0.7150} & \multicolumn{2}{l|}{24.62/0.7280} & \multicolumn{2}{l}{27.90/0.8610} \\
VDSR\cite{kim2016accurate}                    &                        & 666      &  612.6     & \multicolumn{2}{l|}{31.35/0.8838} & \multicolumn{2}{l|}{28.01/0.7674} & \multicolumn{2}{l|}{27.29/0.7251} & \multicolumn{2}{l|}{25.18/0.7524} & \multicolumn{2}{l}{28.83/0.8870} \\
LapSRN\cite{lai2017deep}                      &                        & 813      &  149.4     & \multicolumn{2}{l|}{31.54/0.8852} & \multicolumn{2}{l|}{28.09/0.7700} & \multicolumn{2}{l|}{27.32/0.7275} & \multicolumn{2}{l|}{25.21/0.7562} & \multicolumn{2}{l}{29.09/0.8900} \\
DRRN\cite{tai2017image}                       &                        & 298      &  6,796.9   & \multicolumn{2}{l|}{31.68/0.8888} & \multicolumn{2}{l|}{28.21/0.7720} & \multicolumn{2}{l|}{27.38/0.7284} & \multicolumn{2}{l|}{25.44/0.7638} & \multicolumn{2}{l}{29.45/0.8946} \\
MemNet\cite{tai2017memnet}                    &                        & 678      &  2,662.4   & \multicolumn{2}{l|}{31.74/0.8893} & \multicolumn{2}{l|}{28.26/0.7723} & \multicolumn{2}{l|}{27.40/0.7281} & \multicolumn{2}{l|}{25.50/0.7630} & \multicolumn{2}{l}{29.42/0.8942} \\
IDN\cite{hui2018fast}                         &                        & 553      &  32.3      & \multicolumn{2}{l|}{31.82/0.8903} & \multicolumn{2}{l|}{28.25/0.7730} & \multicolumn{2}{l|}{27.41/0.7297} & \multicolumn{2}{l|}{25.41/0.7632} & \multicolumn{2}{l}{29.41/0.8942} \\
CARN\cite{ahn2018fast}                        &                        & 1592     &  90.9      & \multicolumn{2}{l|}{32.13/0.8937} & \multicolumn{2}{l|}{28.60/0.7806} & \multicolumn{2}{l|}{27.58/0.7349} & \multicolumn{2}{l|}{26.07/0.7837} & \multicolumn{2}{l}{30.47/0.9084} \\
IMDN\cite{hui2019lightweight}                 &                        & 715      &  40.9      & \multicolumn{2}{l|}{32.21/0.8948} & \multicolumn{2}{l|}{28.58/0.7811} & \multicolumn{2}{l|}{27.56/0.7353} & \multicolumn{2}{l|}{26.04/0.7838} & \multicolumn{2}{l}{30.45/0.9075} \\
PAN\cite{zhao2020efficient}                   &                        & 272      &   28.2     & \multicolumn{2}{l|}{32.13/0.8948} & \multicolumn{2}{l|}{28.61/0.7822} & \multicolumn{2}{l|}{27.59/0.7363} & \multicolumn{2}{l|}{26.11/0.7854} & \multicolumn{2}{l}{30.51/\textcolor{blue}{0.9095}} \\
LAPAR-A\cite{li2020lapar}                     &                        & 659      &   94.0     & \multicolumn{2}{l|}{32.15/0.8944} & \multicolumn{2}{l|}{28.61/0.7818} & \multicolumn{2}{l|}{\textcolor{blue}{27.61/0.7366}} & \multicolumn{2}{l|}{\textcolor{blue}{26.14/0.7871}} & \multicolumn{2}{l}{30.42/0.9074} \\
RFDN\cite{rfdn}                    &                        & 550      &  23.9      & \multicolumn{2}{l|}{\textcolor{blue}{32.24/0.8952}} & \multicolumn{2}{l|}{28.61/0.7819} & \multicolumn{2}{l|}{27.57/0.7360} & \multicolumn{2}{l|}{26.11/0.7858} & \multicolumn{2}{l}{\textcolor{blue}{30.58}/0.9089} \\
BSRN-S(Ours)                                  &                        & 156      &  8.3       & \multicolumn{2}{l|}{32.16/0.8949} & \multicolumn{2}{l|}{\textcolor{blue}{28.62/0.7823}} & \multicolumn{2}{l|}{27.58/0.7365} & \multicolumn{2}{l|}{26.08/0.7849} & \multicolumn{2}{l}{30.53/0.9089} \\
BSRN(Ours)                                  &                        & 352      &  19.4      & \multicolumn{2}{l|}{\textcolor{red}{32.35/0.8966}} & \multicolumn{2}{l|}{\textcolor{red}{28.73/0.7847}} & \multicolumn{2}{l|}{\textcolor{red}{27.65/0.7387}} & \multicolumn{2}{l|}{\textcolor{red}{26.27/0.7908}} & \multicolumn{2}{l}{\textcolor{red}{30.84/0.9123}} \\ \bottomrule
\end{tabular}}
\end{table*}

\subsection{Experimental Setup}

\textbf{Datasets and Metrics.}
The training images consist of 2650 images from Flickr2K\cite{lim2017enhanced} and 800 images from DIV2K\cite{div2k}. We use the five standard benchmark datasets of Set5\cite{set5}, Set14\cite{set14}, B100\cite{b100}, Urban100\cite{urban100}, and Manga109\cite{manga109} to evaluate the performance of different approaches. The average peak signal to noise ratio (PSNR) and the structural similarity\cite{ssim} (SSIM) on the Y channel (i.e., luminance) are exploited as the evaluation metrics.

\label{section:implementation_details}
\textbf{Implementation details of BSRN.}
The proposed BSRN consists of 8 ESDBs and the number of channels is set to 64. The kernel size of all depth-wise convolutions is set to 3. Data augmentation methods of random rotation by 90$^{\circ}$, 180$^{\circ}$, 270$^{\circ}$ and flipping horizontally are utilized. The mini-batch size is set to 64 and the patch size of each LR input is set to $48\times48$. The model is trained by Adam optimizer\cite{kingma2014adam} with $\beta_1=0.9, \beta_2=0.999$. The initial learning rate is set to $1\times10^{-3}$ with cosine learning rate decay. $L_1$ loss is used to optimize the model for total $1\times10^{6}$ iterations. We use Pytorch to implement our model on two GeForce RTX 3090 GPUs and the training process costs about 30 hours.

\textbf{Implementation details of BSRN-S for NTIRE2022 Challenge.}
BSRN-S is a small variant of BSRN designed for the challenge, which requires the participants to devise an efficient network while maintaining PSNR of 29.00dB on DIV2K validation dataset. Specifically, we reduce the number of ESDBs to 5 and the number of features to 48. The CCA block is replaced with learnable channel-wise weights. During the training process, the input patch size is set to $64\times64$ and the mini-batch is set to 256. The number of training iterations is increased to $1.5\times10^{6}$ and four GeForce RTX 2080Ti GPUs are used for training.

\begin{figure*}[ht]
  \centering
    \includegraphics[width=0.94\linewidth]{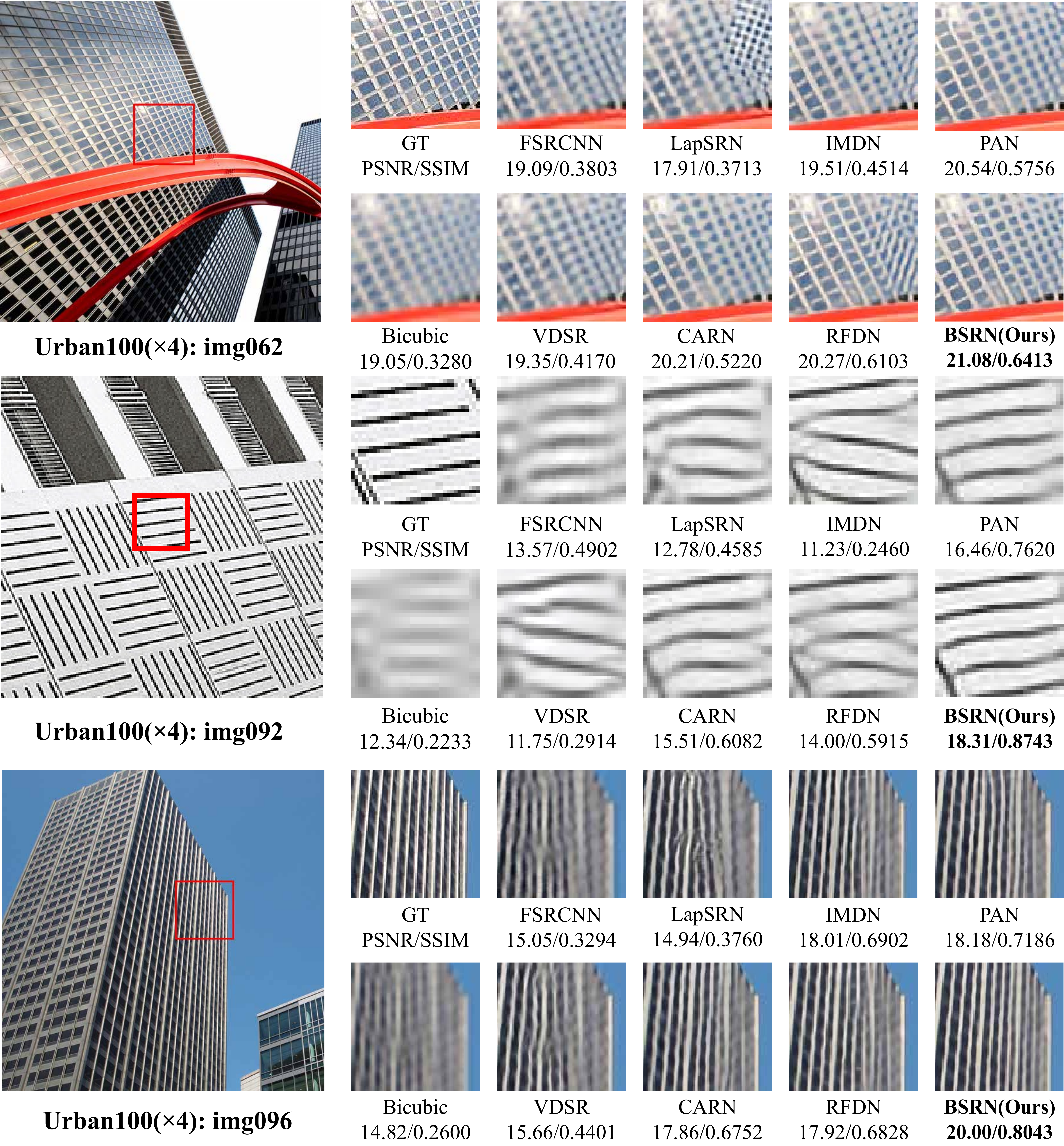}
  \caption{Visual comparison of BSRN with the state-of-the-art methods on $\times$4 SR.}
  \label{fig:visual}
\end{figure*}

\begin{table*}[htbp]
\centering
\caption{Results of NTIRE 2022 Efficient Super-Resolution Sub-Track 1: Model Complexity.}
\label{table:ntire_results}
\resizebox{\linewidth}{!}{%
\begin{tabular}{@{}c|ccccccc@{}}
\toprule
Team                & Val PSNR & Test PSNR  & Params[M] & FLOPs[G] & Acts[M] & Mem[M] & Runtime[ms] \\ \hline
XPixel (Ours)        & 29.01    & 28.69      & 0.156     & 9.496    & 65.76   & 729.94  & 140.47  \\
NJUST\_ESR           & 28.96    & 28.68      & 0.176     & 8.73     & 160.43  & 1346.74 & 164.8\\
HisenseResearch      & 29.00    & 28.72      & 0.242     & 14.51    & 151.36  & 861.84  & 47.75\\
NEESR                & 29.01    & 28.71      & 0.272     & 16.86    & 79.59   & 575.99  & 29.97\\
NKU-ESR              & 29.00    & 28.66      & 0.276     & 16.73    & 111.12  & 662.51  & 34.81\\ 
\hline
RFDN AIM2020 Winner  & 29.04    & 28.75      & 0.433     & 27.1     & 112.03  & 788.13  & 41.97\\
\bottomrule
\end{tabular}%
}
\end{table*}


\subsection{Ablation Study}
In this section, we first present the effects of different convolution decomposition methods. Then we demonstrate the effectiveness of the two attention modules and compare the effects of different activation functions. Finally, we further show the effectiveness of the proposed architecture.

\textbf{Effects of different convolution decompositions.}
We conduct experiments to show the effects of different ways of convolution decomposition based on RFDN. The experimental results are presented in~\cref{table:ablation_conv}. DSConv represents the original depth-wise separable convolution~\cite{howard2017mobilenets}. BSConvU and BSConvS represent the two variants of BSConv proposed in~\cite{bsconv}. We can observe that apparent performance drops appear with significant computation decreases when performing convolution decompositions. Among the three decomposition strategies, BSConvU performs the best, thus we choose to use it in our model.

\textbf{Effectiveness of ESA and CCA.}
We also conduct the ablation study to validate the effectiveness of the two attention modules of ESA and CCA, as depicted in \cref{table:ablation_module}. With about 9\% drop in parameters, an obvious performance drop appears for BSRN without ESA. Compared to BSRN without CCA, the complete BSRN obtains performance gains of 0.5dB on Set5, Urban100 and Manga109 datasets. The results demonstrate that ESA and CCA can effectively enhance the model capacity.

\textbf{Exploration of different activation functions.}
Most of the previous SR networks adopt ReLU~\cite{nair2010rectified} or LeakyReLU~\cite{maas2013rectifier} as the activation function. However, GELU~\cite{hendrycks2016gaussian} is gradually becoming the mainstream choice in recent works. \cite{howard2019searching} investigates the effects of different activation functions in an efficient model MobileNet V3 and proposes a new activation function h-swish. Therefore, we also investigate various activation functions to explore the best choice for our method. The results in \cref{table:activation} show that different activation functions can obviously affect the performance of the model. Among these activation functions, GELU obtains a remarkable performance gain, especially on the Urban100 dataset. Thus, we choose GELU as the activation function in our model.

\textbf{Effectiveness of the proposed architecture.}
We design two variants of BSRN to demonstrate the effectiveness of the proposed architecture. We set the depth and width of BSRN the same as the original RFDN for BSRN-1 and then enlarge the model capacity to the similar computations to RFDN for BSRN-2. Note that we train the compared models under the same training settings for a fair comparison. As shown in \cref{table:BSRN_effectivenes}, we can observe that BSRN-1 outperforms RFDN with less computation. In addition, BSRN-2 obtains a significant performance gain compared to RFDN, especially on the Manga109 dataset. The experimental results show the superiority of the proposed architecture.

\subsection{Comparison with State-of-the-art Methods}
We compare the proposed BSRN with state-of-the-art lightweight SR approaches, including  SRCNN\cite{dong2014learning}, FSRCNN\cite{dong2016accelerating}, VDSR\cite{kim2016accurate}, LapSRN\cite{lai2017deep}, DRRN\cite{tai2017image}, MemNet\cite{tai2017memnet}, IDN\cite{hui2018fast}, CARN\cite{ahn2018fast}, IMDN\cite{hui2019lightweight}, PAN\cite{zhao2020efficient}, LAPAR-A\cite{li2020lapar}, RFDN\cite{rfdn}.
\cref{table:comparison_SOTA} shows the quantitative comparison results for different upscale factors. We also provide the number of parameters and Multi-Adds calculated on the $1280 \times720$ output. Compared to other lightweight SR methods, our BSRN achieves the best performance with only 332K-352K parameters and almost the fewest Multi-Adds. Our solution for NTIRE Challenge, BSRN-S, also obtains competitive performance with only 156K parameters and 8.3G Multi-Adds on $\times$4 SR. The qualitative comparison is demonstrated in~\cref{fig:visual} and our approach can also obtain the best visual quality compared to the state-of-the-art methods.

\subsection{BSRN-S for NTIRE2022 Challenge}
Our BSRN-S won the first place in the NTIRE2022 Efficient Super-Resolution Challenge Sub-Track 1: Model
Complexity, for which the summed rank of the number of parameters and FLOPs are utilized for the final ranking. The results are shown in \cref{table:ntire_results}. Note that we use different settings to train the model for the challenge that is presented in~\cref{section:implementation_details}. For the specific metrics in the table, 'Val PSNR' and 'Test PSNR' are PSNR results tested on the DIV2K validation and test sets. 
%
%
Compared to other competing solutions, our method has the least number of parameters and the second-fewest FLOPs. For the average runtime, it is related to the optimization of the code and the calculation of the specific testing platform for different operators. After optimization, BSRN-S-opt obtains similar runtime to IMDN and RFDN on the same GPU shown in~\cref{table:runtime}.
However, since it is unfriendly for GPU to calculate depth-wise convolution, the runtime of our approach is relatively larger. 


\begin{table}[!htbp]
\centering
\vspace{-3pt}
\caption{Comparison of computational cost.}
\label{table:runtime}
\vspace{-2pt}
\resizebox{\linewidth}{!}{%
\begin{tabular}{@{}c|ccccc@{}}
\toprule
Team            & Params[K]  & Multi-Adds[G]  & Mem[M]  & Runtime[ms] \\ \hline
MSRResNet       & 1517.57    & 166.36   & 598.55  & 70.83  \\
IMDN            & 893.94     & 58.53    & 120.17  & 25.32  \\
RFDN            & 433.45     & 27.10    & 201.59  & 26.53  \\
BSRN-S(ours)    & 156.05     & 8.35     & 184.57  & 36.51  \\
BSRN-S-opt(ours)    & 156.05     & 8.35     & 184.57  & 26.81  \\
\bottomrule 
\end{tabular}%
}
\end{table}
\vspace{-10pt}

\section{Conclusions}
In this paper, we propose a lightweight network for single image super-resolution called the blueprint separable residual network (BSRN). The design of BSRN is inspired by the residual feature distillation network (RFDN) and the blueprint separable convolution (BSConv). We adopt the similar architecture of RFDN but introduce a more efficient blueprint shallow residual block (BSRB) by replacing the standard convolution with BSConv in the shallow residual block (SRB) in RFDN. Moreover, we use the effective ECA block and CCA block to enhance the representative ability of the model. Extensive experiments show that our method achieves the best performance with fewer parameters and Multi-Adds compared to the state-of-the-art efficient SR methods. Besides, our solution won first place in the model complexity track of the NTIRE 2022 efficient super-resolution challenge.

\vspace{-5pt}
\paragraph{Acknowledgements.} This work is partially supported by the National Natural Science Foundation of China (61906184), the Joint Lab of CAS-HK, the Shenzhen Research Program(RCJC20200714114557087), the Shanghai Committee of Science and Technology, China (Grant No. 21DZ1100100).

\clearpage
\clearpage
{\small
\bibliographystyle{ieee_fullname}
\bibliography{egbib}
}

\end{document}